\documentclass[letterpaper]{article} 
\usepackage{aaai23}  
\usepackage{times}  
\usepackage{helvet}  
\usepackage{courier}  
\usepackage[hyphens]{url}  
\usepackage{graphicx} 
\usepackage{natbib}  
\usepackage{caption} 
\usepackage{amsmath,amsfonts,amssymb}
\usepackage{algorithm}
\usepackage{algorithmic}
\usepackage{booktabs}
\usepackage{newfloat}
\usepackage{listings}
\DeclareCaptionStyle{ruled}{labelfont=normalfont,labelsep=colon,strut=off} 
\floatstyle{ruled}
\newfloat{listing}{tb}{lst}{}
\floatname{listing}{Listing}
\title{Device Tuning for Multi-Task Large Model}
\author{
    Penghao Jiang\textsuperscript{\rm 1}\thanks{These authors contributed equally.}, Xuanchen Hou\textsuperscript{\rm 2*}, Yinsi Zhou\textsuperscript{\rm 1}
}
\affiliations{
    \textsuperscript{\rm 1}The Australia National University\\
    \textsuperscript{\rm 2}Deakin university\\

}

\usepackage{bibentry}

\begin{document}

\maketitle

\begin{abstract}
Unsupervised pre-training approaches have achieved
great success in many fields such as Computer Vision
(CV), Natural Language Processing (NLP) and so on.
However, compared to typical deep learning models,
pre-training or even fine-tuning the state-of-the-art selfattention
models is extremely expensive, as they require
much more computational and memory resources. It
severely limits their applications and success in a variety
of domains, especially for multi-task learning. To
improve the efficiency, we propose Device Tuning for efficient
multi-task model, which is massively multi-task
framework across cloud and device, and is designed to encourage
learning of representations that generalize better
to many different tasks. Specifically, we design Device
Tuning architecture of multi-task model that benefit both
cloud modeling and device modeling, which reduces the
communication between device and cloud by representation
compression. Experimental results demonstrate the
effectiveness of our proposed method.
\end{abstract}

\section{Introduction}
Self-attention-based models, especially vision transformers
\citep{5}, are an alternative to convolutional
neural networks (CNNs) to learn visual representations.
Briefly, ViT divides an image into a sequence of
non-overlapping patches and then learns inter-patch representations
using multi-headed self-attention in transformers
\citep{12}. The general trend is to
increase the number of parameters in ViT networks to
improve the performance (e.g., \citet*{14,7,18}). However, these
performance improvements come at the cost of model
size (network parameters) and latency. Many real-world
applications (e.g., augmented reality and autonomous
wheelchairs) require visual recognition tasks (e.g., object
detection and semantic segmentation) to run on resourceconstrained
mobile devices in a timely fashion. To be effective, ViT models for such tasks should be lightweight
and fast. Even if the model size of ViT models
is reduced to match the resource constraints of mobile
devices, their performance is significantly worse than
light-weight CNNs. For instance, for a parameter budget
of about 5-6 million, DeIT \citep{4} is 3%
less accurate than MobileNetv3 \citep{10}.
However, it is still extremely expensive to pretrain or
even just to fine-tune the Transformer layers, as they require
much more computational and memory resources
compared to traditional models. This largely limits their
applications and success in more fields.

To reduce the computational and memory resources for
cloud centralized model, recent works \citep{6}
explored a split deployment across cloud and device,
which could reduce the inference cost and memory resources.
Such works about mobile computing and the
Internet of Things (IoTs) are driving computing toward
dispersion. The increasing capacity of mobile devices
makes it possible to consider the intelligence services,
such as online machine translation and online dialogue
modeling, from cloud to device modeling. Several recent
works in different perspectives like privacy \citep{1}, efficiency \citep{8}, applications \citep{6} have explored
this pervasive computing advantages. There have
been some efforts to distill BERT into resource-limited
mobile devices. However, how to leverage the advantages
of the device modeling and the cloud modeling
jointly to benefit both sides is still a challenge for unsupervised
pre-training models.

The first issue of this challenge is how to design an architecture
that not only has a lower resource-to-performance
ratio on device but also take advantage of the device modeling
and the cloud modeling jointly. The second issue
of this challenge is how to design an effective multi-task
framework which could learn one general scalable and
lighter model.

\begin{figure}[t]
	\centering
	\includegraphics[width=\columnwidth]{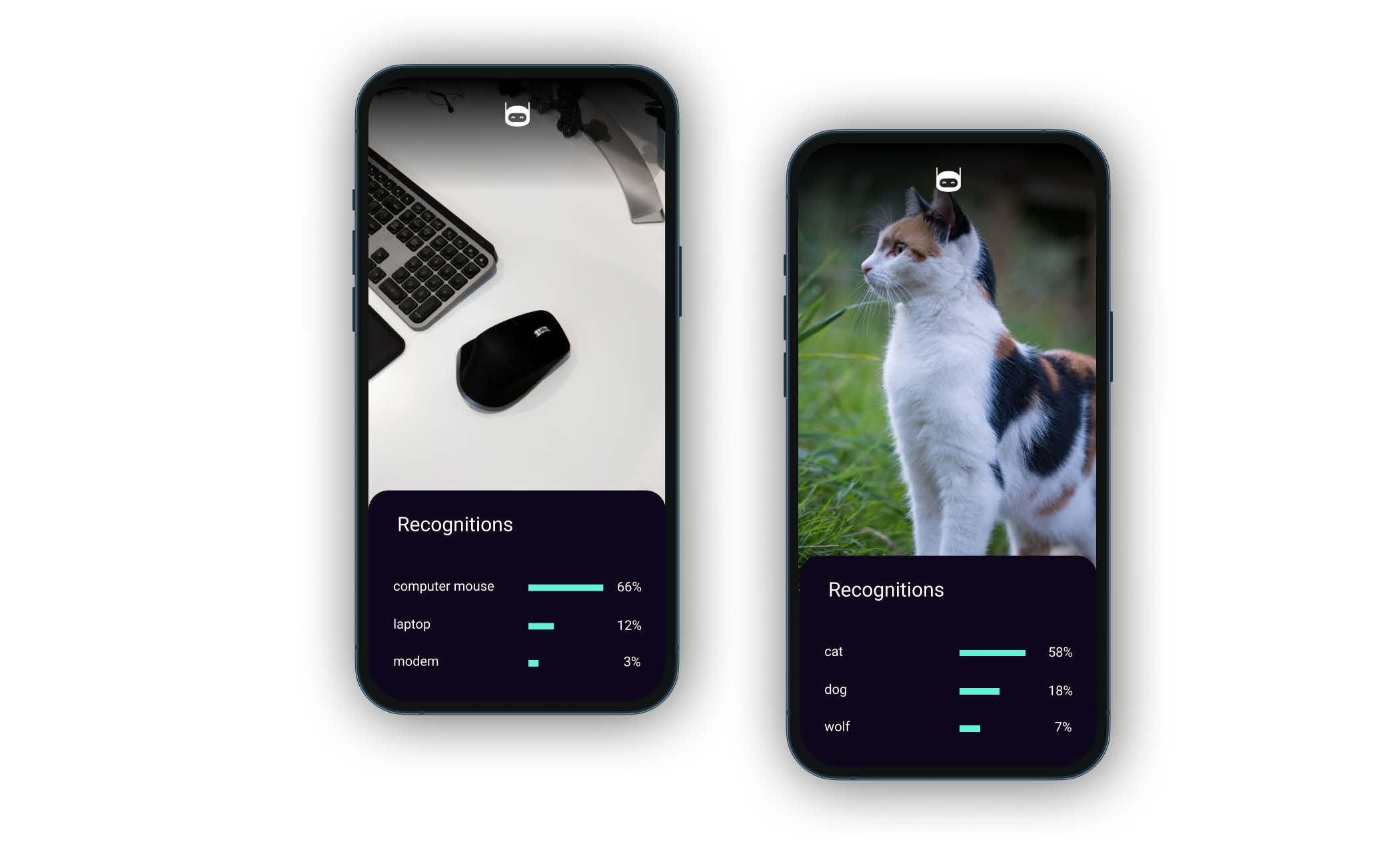} 
	\caption{Image recognitions on mobiles.}
	\label{fig:1}
\end{figure}

To overcome challenges mentioned above, we propose
Device Tuning framework, which is one general framework
across cloud and device for multiple tasks. As shown in Figure 1, previous unsupervised pre-training
methods learn a centralized could model, models designed
for resource-limited mobile devices learn a task
specific device model. Different from these methods,
our device tuning method share parameters in could and
learn task specific parameters in device. Specifically, to
overcome the first issue, we propose a general framework
including device encoder and cloud decoder, which reduces
the communication by representation compression.
Then, to overcome the second issue, we consider a gradient
normalization method which automatically balances
training in multi-task framework by dynamically tuning
gradient magnitudes. In summary, the contributions of
this paper are:
\begin{itemize}
\item Different from existing works that either only consider
the cloud modeling, or on-device modeling,
we design Device Tuning architecture of multi-task
model that benefit both cloud modeling and device
modeling.
\item We consider a novel method which reduces the communication
between device and cloud by representation
compression.
\item Extensive experiments show that our proposed Device
tuning framework can significantly improve
methods in different tasks.
\end{itemize}

\subsection{RelatedWork}
\citet{5} apply transformers of \citet{16} for large-scale image recognition and
showed that with extremely large-scale datasets (e.g.,
JFT-300M), ViTs can achieve CNN-level accuracy without
image-specific inductive bias. With extensive data
augmentation, heavy L2 regularization, and distillation,
ViTs can be trained on the ImageNet dataset to achieve
CNN-level performance \cite{14,15,21}. However, unlike CNNs, ViTs show substandard
optimizability and are difficult to train. Subsequent
works (e.g., \citet*{7,3,11,17,20,2}) shows that this substandard optimizability
is due to the lack of spatial inductive biases in ViTs.
Incorporating such biases using convolutions in ViTs
improves their stability and performance. Different designs
have been explored to reap the benefits of convolutions
and transformers. For instance, ViT-C of \citet{19} adds an early convolutional stem to ViT.
CvT \citep{18} modifies the multi-head attention
in transformers and uses depth-wise separable convolutions
instead of linear projections. BoTNet \citep{12} replaces the standard 3 $\times$ 3 convolution
in the bottleneck unit of ResNet with multi-head attention.
ConViT \citep{4} incorporates soft
convolutional inductive biases using a gated positional
self-attention. PiT \citep{9} extends ViT with
depth-wise convolution-based pooling layer. Though
these models can achieve competitive performance to
CNNs with extensive augmentation, the majority of these
models are heavy-weight. For instance, PiT and CvT
learns 6.1 $\times$ and 1.7 $\times$ more parameters than Efficient-Net \citep{13} and achieves similar performance
(top-1 accuracy of about 81.6\%) on ImageNet-1k dataset,
respectively. Also, when these models are scaled down
to build light-weight ViT models, their performance is
significantly worse than light-weight CNNs. For a parameter
budget of about 6 million, ImageNet-1k accuracy of
PiT is 2.2\% less than MobileNetv3.

\section{Method}
\subsection{Preliminary}
\subsubsection{Transformer}
Transformer layers (Vaswani et al. 2017) have achieved state-of-the-art performance across various tasks, which is a highly modularized neural network. Each Transformer layer consists of two sub-modules: multi-head self-attention (S-Attn) and position-wise feed-forward network (P-FFN). A residual connection and layer normalization wrap both sub-modules. The computation of a single Transformer layer with a length $T$ sequence of hidden states $\mathbf{h}=\left[h_1, \ldots, h_T\right]$ can be expressed as
\begin{align}
\mathbf{h} & \leftarrow \operatorname{LayerNorm}(\mathbf{h}+\mathrm{S}-\operatorname{Attn}(\mathrm{Q}, \mathrm{K}, \mathrm{V}=\mathbf{h})) \label{eq1}\\
h_i & \leftarrow \operatorname{LayerNorm}\left(h_i+\operatorname{P-FFN}\left(h_i\right)\right), \quad \forall i=1, \cdots, T\label{eq2}
\end{align}

\subsection{Device Tuning}
To design a general and efficient framework across cloud
and device for multiple tasks, the main challenge is to
solve the computational efficiency problem and reduce
the communication. To achieve representation compression
and computation reduction, our model employs a
device encoder that reduces the sequence length of the
hidden states, which keeps the same overall skeleton of
interleaved multi-head self-attention and position-wise
feed-forward network and inheriting the high capacity and optimization advantages of the Transformer architecture.

\begin{figure}[t]
	\centering
	\includegraphics[width=\columnwidth]{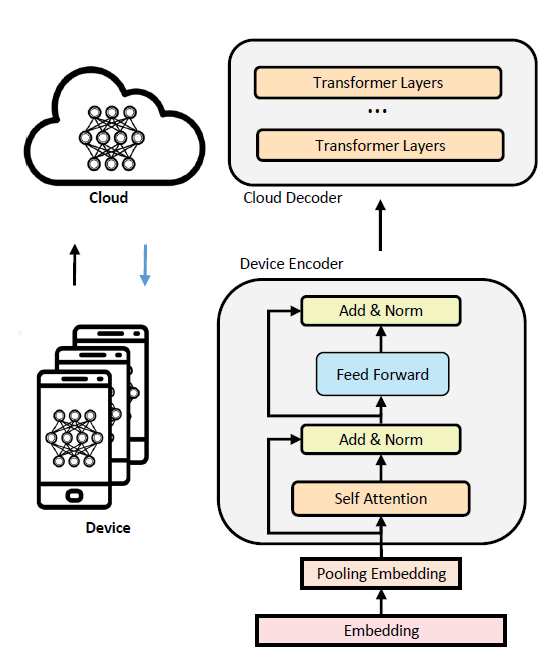} 
	\caption{An overview of our Device Tuning model.}
	\label{fig:2}
\end{figure}

To solve the computational efficiency problem and reduce the communication, we consider a device encoder that reduces the sequence length of the hidden states as shown in Figure 2. Device encoder reduces the length of the hidden sequence by performing a certain type of pooling along the sequence dimension. For hidden sequence $h$, we have $\mathbf{h}^{\prime} \leftarrow \operatorname{Pooling}(\mathbf{h})$, where $\mathbf{h} \in \mathbb{R}^{T \times D}$ and $\mathbf{h}^{\prime} \in \mathbb{R}^{T^{\prime} \times D}$ for some $T^{\prime}<T$. Thus, the query, key and value vectors in self-attention layer are
\begin{equation}
\mathbf{h} \leftarrow \text { LayerNorm }\left(\mathbf{h}^{\prime}+\mathrm{S}-\operatorname{Attn}\left(\mathrm{Q}, \mathrm{K}, \mathrm{V}=\mathbf{h}^{\prime}\right)\right)
\end{equation}
It is worth noting that this multi-head self-attention (SAttn module) module's output sequence is the same length as the pooled sequence $\mathbf{h}^{\prime}$. Such pooling strategy merge (or compress) close tokens into a larger semantic component, which intuitively follows the linguistic prior. The rest of the encoder computation just follows the typical updates in Eq.\eqref{eq2} and \eqref{eq1} once the sequence length is halved following the pooling attention. The output of device encoder is given to cloud decoder.

\section{Experiments}
In this section, we conduct experiments on benchmarks
to evaluate the effectiveness of the proposed frameworks
by first pretraining it and then fine-tuning it in downstream
tasks.

\subsection{Performance Comparison}
\subsubsection{Same-scale Results}
In same-scale, we compare device
tuning to the standard Transformer models with similar amount of computation. We choose recent models with
similar paraters as ours. The results are shown in Table
\ref{tab1}. From Table \ref{tab1}, we could find that our proposed method
outperform baselines in all cases, which demonstrate the
effectiveness of proposed method.

\subsection{Different-scale Results}
To show the effectiveness of
our proposed Device Tuning, we compare our Device
Tuning with different backbones. The results are shown
in Table \ref{tab2}. The models are trained in the same settings.
Similar to the similar-scale results, our method outperforms
baselines in all cases, suggesting the good scalability
of our proposed Device Tuning.

\begin{table}[t]
	\centering
	\caption{Compare with similar parameters' networks on
		ImageNet-1k.}
	\label{tab1}
	\begin{tabular}{lrr}
		\toprule[2pt] Model & \# Params. $\Downarrow$ & Top-1 $\Uparrow$ \\\midrule
		MobileNetv1 & 2.6 {M} & 68.4 \\
		MobileNetv2 & 2.6 {M} & 69.8 \\
		MobileNetv3 & 2.5 {M} & 67.4 \\
		ShuffleNetv2 & 2.3 {M} & 69.4 \\
		ESPNetv2 & 2.3 {M} & 69.2 \\
		Ours & 2.2 {M} & \textbf{70.6} \\
		\bottomrule
	\end{tabular}
\end{table}

\begin{table}[t]
	\centering
	\caption{with different backbones on ImageNet-
		1k.}
	\label{tab2}
	\begin{tabular}{lrr}
		\toprule[2pt] Model & \# Params. $\Downarrow$ & Top-1 $\Uparrow$ \\\midrule
			 { DenseNet-169 } & 14 {M} & 76.2 \\
			 { EfficientNet-B0 } & 5.3 {M} & 76.3 \\
			 { ResNet-101 } & 44.5 {M} & 77.4 \\
			 { ResNet-101-SE } & 49.3 {M} & 77.6 \\
			 { Ours } & 15.6 {M} & \textbf{78.0}\\
		\bottomrule
	\end{tabular}
\end{table}

\section{Conclusion}
Recently, unsupervised pre-training methods have
achieved great success in many fields such as Computer
Vision (CV), Natural Language Processing (NLP) and
so on. However, it is extremely expensive to pretrain or
even just to fine-tune the state-of-the-art self-attention
models, as they require much more FLOPs and memory
resources compared to traditional models. To improve
the efficiency, we propose Device Tuning for efficient
multi-task model, which is massively multi-task framework
across cloud and device, and is designed to encourage
learning of representations that generalize better to
many different tasks. Specifically, we design an architecture
that not only has a lower resource-to-performance
ratio on device but also take advantages of the device
modeling and the cloud modeling jointly. Experimental
results demonstrate the effectiveness of our proposed
method.

\bibliography{Device}

\begin{thebibliography}{21}
\providecommand{\natexlab}[1]{#1}

\bibitem[{Bistritz, Mann, and Bambos(2020)}]{1}
Bistritz, I.; Mann, A.; and Bambos, N. 2020.
\newblock Distributed distillation for on-device learning.
\newblock \emph{Advances in Neural Information Processing Systems}, 33:
  22593--22604.

\bibitem[{Chen et~al.(2021)Chen, Dai, Chen, Liu, Dong, Yuan, and Liu}]{2}
Chen, Y.; Dai, X.; Chen, D.; Liu, M.; Dong, X.; Yuan, L.; and Liu, Z. 2021.
\newblock Mobile-former: Bridging mobilenet and transformer.
\newblock In \emph{Proceedings of the IEEE/CVF Conference on Computer Vision
  and Pattern Recognition}, 5270--5279.

\bibitem[{Dai et~al.(2021)Dai, Liu, Le, and Tan}]{3}
Dai, Z.; Liu, H.; Le, Q.~V.; and Tan, M. 2021.
\newblock Coatnet: Marrying convolution and attention for all data sizes.
\newblock \emph{Advances in Neural Information Processing Systems}, 34:
  3965--3977.

\bibitem[{d'Ascoli et~al.(2021)d'Ascoli, Touvron, Leavitt, Morcos, Biroli, and
  Sagun}]{4}
d'Ascoli, S.; Touvron, H.; Leavitt, M.~L.; Morcos, A.~S.; Biroli, G.; and
  Sagun, L. 2021.
\newblock Convit: Improving vision transformers with soft convolutional
  inductive biases.
\newblock In \emph{International Conference on Machine Learning}, 2286--2296.
  PMLR.

\bibitem[{Dosovitskiy et~al.(2021)Dosovitskiy, Beyer, Kolesnikov, Weissenborn,
  Zhai, Unterthiner, Dehghani, Minderer, Heigold, Gelly et~al.}]{5}
Dosovitskiy, A.; Beyer, L.; Kolesnikov, A.; Weissenborn, D.; Zhai, X.;
  Unterthiner, T.; Dehghani, M.; Minderer, M.; Heigold, G.; Gelly, S.; et~al.
  2021.
\newblock An image is worth 16x16 words: Transformers for image recognition at
  scale.
\newblock \emph{arXiv preprint arXiv:2010.11929}.

\bibitem[{Gong et~al.(2020)Gong, Jiang, Feng, Hu, Zhao, Liu, and Ou}]{6}
Gong, Y.; Jiang, Z.; Feng, Y.; Hu, B.; Zhao, K.; Liu, Q.; and Ou, W. 2020.
\newblock EdgeRec: recommender system on edge in Mobile Taobao.
\newblock In \emph{Proceedings of the 29th ACM International Conference on
  Information \& Knowledge Management}, 2477--2484.

\bibitem[{Graham et~al.(2021)Graham, El-Nouby, Touvron, Stock, Joulin,
  J{\'e}gou, and Douze}]{7}
Graham, B.; El-Nouby, A.; Touvron, H.; Stock, P.; Joulin, A.; J{\'e}gou, H.;
  and Douze, M. 2021.
\newblock Levit: a vision transformer in convnet's clothing for faster
  inference.
\newblock In \emph{Proceedings of the IEEE/CVF international conference on
  computer vision}, 12259--12269.

\bibitem[{Han, Mao, and Dally(2016)}]{8}
Han, S.; Mao, H.; and Dally, W.~J. 2016.
\newblock Deep Compression: Compressing Deep Neural Network with Pruning
  Trained Quantization and Huffman Coding.
\newblock In \emph{International Conference on Learning Representations}.

\bibitem[{Heo et~al.(2021)Heo, Yun, Han, Chun, Choe, and Oh}]{9}
Heo, B.; Yun, S.; Han, D.; Chun, S.; Choe, J.; and Oh, S.~J. 2021.
\newblock Rethinking spatial dimensions of vision transformers.
\newblock In \emph{Proceedings of the IEEE/CVF International Conference on
  Computer Vision}, 11936--11945.

\bibitem[{Howard et~al.(2019)Howard, Sandler, Chu, Chen, Chen, Tan, Wang, Zhu,
  Pang, Vasudevan et~al.}]{10}
Howard, A.; Sandler, M.; Chu, G.; Chen, L.-C.; Chen, B.; Tan, M.; Wang, W.;
  Zhu, Y.; Pang, R.; Vasudevan, V.; et~al. 2019.
\newblock Searching for mobilenetv3.
\newblock In \emph{Proceedings of the IEEE/CVF international conference on
  computer vision}, 1314--1324.

\bibitem[{Liu et~al.(2021)Liu, Lin, Cao, Hu, Wei, Zhang, Lin, and Guo}]{11}
Liu, Z.; Lin, Y.; Cao, Y.; Hu, H.; Wei, Y.; Zhang, Z.; Lin, S.; and Guo, B.
  2021.
\newblock Swin transformer: Hierarchical vision transformer using shifted
  windows.
\newblock In \emph{Proceedings of the IEEE/CVF international conference on
  computer vision}, 10012--10022.

\bibitem[{Srinivas et~al.(2021)Srinivas, Lin, Parmar, Shlens, Abbeel, and
  Vaswani}]{12}
Srinivas, A.; Lin, T.-Y.; Parmar, N.; Shlens, J.; Abbeel, P.; and Vaswani, A.
  2021.
\newblock Bottleneck transformers for visual recognition.
\newblock In \emph{Proceedings of the IEEE/CVF conference on computer vision
  and pattern recognition}, 16519--16529.

\bibitem[{Tan and Le(2019)}]{13}
Tan, M.; and Le, Q. 2019.
\newblock Efficientnet: Rethinking model scaling for convolutional neural
  networks.
\newblock In \emph{International conference on machine learning}, 6105--6114.
  PMLR.

\bibitem[{Touvron et~al.(2021{\natexlab{a}})Touvron, Cord, Douze, Massa,
  Sablayrolles, and J{\'e}gou}]{14}
Touvron, H.; Cord, M.; Douze, M.; Massa, F.; Sablayrolles, A.; and J{\'e}gou,
  H. 2021{\natexlab{a}}.
\newblock Training data-efficient image transformers \& distillation through
  attention.
\newblock In \emph{International conference on machine learning}, 10347--10357.
  PMLR.

\bibitem[{Touvron et~al.(2021{\natexlab{b}})Touvron, Cord, Sablayrolles,
  Synnaeve, and J{\'e}gou}]{15}
Touvron, H.; Cord, M.; Sablayrolles, A.; Synnaeve, G.; and J{\'e}gou, H.
  2021{\natexlab{b}}.
\newblock Going deeper with image transformers.
\newblock In \emph{Proceedings of the IEEE/CVF International Conference on
  Computer Vision}, 32--42.

\bibitem[{Vaswani et~al.(2017)Vaswani, Shazeer, Parmar, Uszkoreit, Jones,
  Gomez, Kaiser, and Polosukhin}]{16}
Vaswani, A.; Shazeer, N.; Parmar, N.; Uszkoreit, J.; Jones, L.; Gomez, A.~N.;
  Kaiser, {\L}.; and Polosukhin, I. 2017.
\newblock Attention is all you need.
\newblock \emph{Advances in neural information processing systems}, 30.

\bibitem[{Wang et~al.(2021)Wang, Xie, Li, Fan, Song, Liang, Lu, Luo, and
  Shao}]{17}
Wang, W.; Xie, E.; Li, X.; Fan, D.-P.; Song, K.; Liang, D.; Lu, T.; Luo, P.;
  and Shao, L. 2021.
\newblock Pyramid vision transformer: A versatile backbone for dense prediction
  without convolutions.
\newblock In \emph{Proceedings of the IEEE/CVF international conference on
  computer vision}, 568--578.

\bibitem[{Wu et~al.(2021)Wu, Xiao, Codella, Liu, Dai, Yuan, and Zhang}]{18}
Wu, H.; Xiao, B.; Codella, N.; Liu, M.; Dai, X.; Yuan, L.; and Zhang, L. 2021.
\newblock Cvt: Introducing convolutions to vision transformers.
\newblock In \emph{Proceedings of the IEEE/CVF International Conference on
  Computer Vision}, 22--31.

\bibitem[{Xiao et~al.(2021)Xiao, Singh, Mintun, Darrell, Doll{\'a}r, and
  Girshick}]{19}
Xiao, T.; Singh, M.; Mintun, E.; Darrell, T.; Doll{\'a}r, P.; and Girshick, R.
  2021.
\newblock Early convolutions help transformers see better.
\newblock \emph{Advances in Neural Information Processing Systems}, 34:
  30392--30400.

\bibitem[{Yuan et~al.(2021)Yuan, Chen, Wang, Yu, Shi, Jiang, Tay, Feng, and
  Yan}]{20}
Yuan, L.; Chen, Y.; Wang, T.; Yu, W.; Shi, Y.; Jiang, Z.-H.; Tay, F.~E.; Feng,
  J.; and Yan, S. 2021.
\newblock Tokens-to-token vit: Training vision transformers from scratch on
  imagenet.
\newblock In \emph{Proceedings of the IEEE/CVF international conference on
  computer vision}, 558--567.

\bibitem[{Zhou et~al.(2021)Zhou, Kang, Jin, Yang, Lian, Jiang, Hou, and
  Feng}]{21}
Zhou, D.; Kang, B.; Jin, X.; Yang, L.; Lian, X.; Jiang, Z.; Hou, Q.; and Feng,
  J. 2021.
\newblock Deepvit: Towards deeper vision transformer.
\newblock \emph{arXiv preprint arXiv:2103.11886}.

\end{thebibliography}

\end{document}